\documentclass{article}

\usepackage[final]{neurips_2024}

\usepackage[utf8]{inputenc} 
\usepackage[T1]{fontenc}    
\usepackage{hyperref}       
\usepackage{url}            
\usepackage{booktabs}       
\usepackage{amsfonts}       
\usepackage{nicefrac}       
\usepackage{microtype}      
\usepackage{xcolor}         

\usepackage{lipsum}
\usepackage{graphicx}
\usepackage{pgf-pie}

\usepackage{multirow}
\usepackage{adjustbox}

\title{MERaLiON: Enhancing Cross-Lingual Understanding of Chinese, Indonesian, Malay and Singlish in Language Models}

\title{MERaLiON-TextLLM: \\ Cross-Lingual Understanding of Large Language Models in Chinese, Indonesian, Malay, and Singlish
}

\author{%
    {\Large MERaLiON Team}
    \AND
    Xin Huang,\quad Tarun Kumar Vangani,\quad  Minh Duc Pham,\quad Xunlong Zou \vspace{1mm} \\ 
    \textbf{Bin Wang,\quad~Zhengyuan Liu,\quad~Ai Ti Aw}
  \vspace{2mm}\\
  Institute for Infocomm Research (I$^2$R), A*STAR, Singapore
}

\begin{document}

\maketitle

\begin{abstract}
    
    Multilingual large language models (MLLMs) have shown impressive capabilities across a variety of languages. However, efficacy can differ greatly between different language families, especially for those with limited linguistic resources. 
    This report presents \textbf{MERaLiON-TextLLM}, a series of open-source language models specifically tailored to improve understanding and generation in Chinese, Indonesian, Malay, and Singlish. The initial released model is built on Llama-3-8B-Base and refined through a meticulously crafted process of continued pre-training and weight merging. 
    Our approach achieves performance improvements across benchmarks in these languages, exceeding the capabilities of the official Llama-3 models. We provide the model checkpoints as a resource to support further research and development in cross-lingual language understanding.\footnote{
    Please cite this report as authored by MERaLiON team.\\
    Corresponding: \{huangx2, vangani\_tarun\_kumar\}@i2r.a-star.edu.sg
    \\
    Available at: \url{https://huggingface.co/MERaLiON/LLaMA-3-MERaLiON-8B-Instruct}
  }

\end{abstract}

\section{Introduction}

Our first released model is named LLaMA-3-MERaLiON-8B-Instruct. LLaMA-3-MERaLiON-8B-Instruct has been extensively pre-trained on English, Chinese, and Indonesian, building upon LLaMA-3.1-8B-Base \cite{grattafiori2024llama3herdmodels}, with a primary emphasis on enhancing its understanding and generation capabilities in Chinese and Indonesian. Leveraging innovative corpus mixing strategies tailored to multilingual regional datasets, we carefully diversified the training materials using domain classification, hyperparameter optimization, and strategic replay techniques. These methods are specifically designed to prevent catastrophic forgetting, enabling the model to retain previously acquired knowledge while significantly improving its ability to produce high-quality, contextually accurate responses in Southeast Asian languages.


\section{Pre-Training}

LLaMA-3-MERaLiON-8B-Instruct training was conducted using the MaxText \cite{maxtext2024} platform, utilizing both NVIDIA H100 GPUs and TPU v4-128 chips. Specifically, we utilized 64 H100 GPUs, achieving approximately 400 TFLOPS per GPU, and TPU v4-128 configurations, attaining around 168 TFLOPS per TPU chip. 

These performance metrics were achieved through optimized sharding, checkpoint strategies, and the selection of optimal batch sizes, to ensure efficient and effective model training. We conducted multiple runs to achieve optimal performance and sampling strategies. 


The pre-training data was allocated across three primary languages: English, Indonesian, and Chinese. Specifically, 38 billion tokens were allocated to English, 45 billion to Indonesian, and 42 billion to Chinese. This allocation is designed to ensure robust performance across all three languages, catering to a variety of linguistic and cultural contexts. 

\begin{figure}[t]
    \centering
    \begin{tikzpicture}
        \pie[
            text=legend, 
            radius=3 
        ]{
            30.4/English (38\,B), 
            36/Indonesian (45\,B), 
            33.6/Chinese (42\,B)
        }
    \end{tikzpicture}
    \caption{Distribution of Tokens}
\label{tab:tokens-dist}
\end{figure}
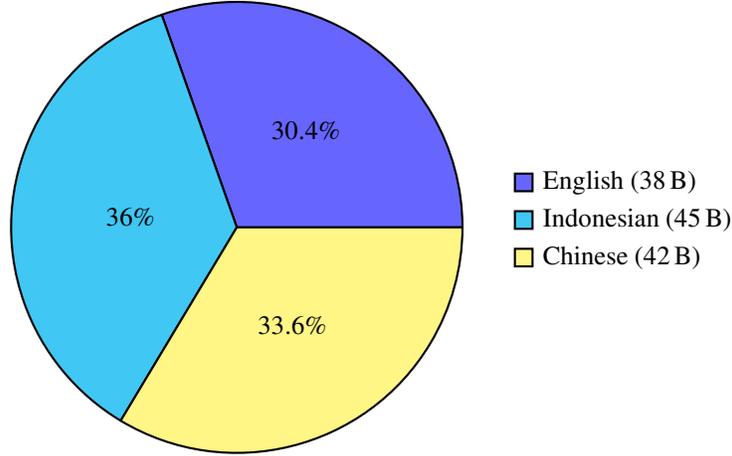

Figure \ref{tab:tokens-dist} illustrates the token distribution in a pie chart format for clear visualization. English accounted for 30.4\% of the total tokens, Indonesian for 36.0\%, and Chinese for 33.6\%. The allocation highlights the model's multilingual focus, ensuring no language dominates the training set, thereby fostering balanced performance.

\section{Instruction Tuning and Model Merging}

\begin{table}[b]
\centering
\caption{Performance comparison of the merged model (MERaLiON-LLaMA-3.1-8B-Instruct) against baseline and instruction-tuned models on Cross-MMLU and Cross-LogiQA benchmarks.}
\label{tab:cross-combined}
\begin{adjustbox}{width=1.00\textwidth,center}
\begin{tabular}{|l|ccc|ccc|}
\hline
\multirow{2}{*}{\textbf{Model}} & \multicolumn{3}{c|}{\textbf{Cross-MMLU}} & \multicolumn{3}{c|}{\textbf{Cross-LogiQA}} \\
\cline{2-7}
 & English & Chinese & Indonesian & English & Chinese & Indonesian \\
\hline
\textbf{LLaMA-3-MERaLiON-8B-Instruct} & \textbf{0.85} & \textbf{0.69} & \textbf{0.71} & \textbf{0.591} & 0.528 & \textbf{0.494} \\  \hline
Meta-Llama-3.1-8B-Instruct             & 0.82 & 0.63 & 0.66 & 0.585 & \textbf{0.585} & 0.455 \\ \hline
Instruction-Tuned Llama-3.1-8B         & 0.73 & 0.60 & 0.58 & 0.563 & 0.523 & 0.466 \\ 
\hline
\end{tabular}
\end{adjustbox}
\end{table}

We present updated experimental results for our latest variant of the Southeast Asian language model, focusing on three representative languages: English, Chinese, and Indonesian. Our work began with the construction of a multilingual instruction-response corpus containing approximately 3M pairs. Using this dataset, we performed instruction tuning on the Llama-3.1-8B base model to evaluate potential performance gains. However, our initial experiments revealed that this approach alone did not surpass the robust baseline performance of the Llama-3.1-8B-Instruct model. Specifically, the instruction-tuned model underperformed the original Llama-3.1-8B-Instruct across the three target languages in the Cross-MMLU and Cross-LogiQA datasets~\cite{wang-etal-2024-seaeval}.

To address these limitations, we explored model merging techniques aimed at enhancing instruction-following capabilities without requiring extensive additional tuning. By carefully merging the weight sets of Llama-3.1-8B-Base and Llama-3.1-8B-Instruct, we leveraged the strong instruction-following performance of the instructed variant while integrating domain-specific knowledge from our custom dataset. As illustrated in Table~\ref{tab:cross-combined}, the resulting "MERaLiON-Llama-3.1-8B-Instruct" model consistently outperformed both the baseline Llama-3.1-8B-Instruct and our standalone instruction-tuned variant across English, Chinese, and Indonesian test sets.

These results underscore the effectiveness of merging pretrained instructed weights with domain-specific models to achieve enhanced instruction-following capabilities. The MERaLiON-Llama-3.1-8B-Instruct model demonstrates improved performance in all evaluated languages, confirming the viability of this approach. This hybrid methodology offers a promising avenue for the development of resource-efficient, multilingual, and domain-adaptive instruction-following models without necessitating extensive iterative tuning on large-scale, specialized corpora.




\section{Benchmarks and Evaluation}

\begin{table}[h]
\centering
\caption{Cross-MMLU}
\label{tab:cross-mmlu}
\begin{adjustbox}{width=1.00\textwidth,center}
\begin{tabular}{|l|l|cccc|c|}
\hline
\textbf{Model Series} & \textbf{Model} & \textbf{English} & \textbf{Chinese} & \textbf{Indonesian} & \textbf{Malay} & \textbf{Avg } \\
\hline
\multirow{4}{*}{LLaMA3} & \textbf{LLaMA-3-MERaLiON-8B-Instruct} & \textbf{0.847} & \textbf{0.693} & \textbf{0.713} & 0.613 & \textbf{0.717} \\ \cline{2-7}
& Meta-Llama-3.1-8B-Instruct & 0.82 & 0.633 & 0.66 & \textbf{0.647} & 0.690 \\ \cline{2-7}
& Llama3-8B-SEA-LION-v2.1-Instruct \cite{sea_lion_2024} & 0.753 & 0.667 & 0.693 & 0.64 & 0.688 \\ \cline{2-7}
& Meta-Llama-3-8B-Instruct & 0.767 & 0.653 & 0.573 & 0.573 & 0.642 \\ 
\hline
\multirow{5}{*}{Others} & \textbf{GPT4o-0513} & \textbf{0.927} & \textbf{0.887} & \textbf{0.88} & \textbf{0.907} & \textbf{0.900} \\ \cline{2-7}
& Gemma-2-9B-IT \cite{gemmateam2024gemma2improvingopen} & 0.84 & 0.793 & 0.78 & 0.747 & 0.790 \\ \cline{2-7}
& Gemma2-9B-SEA-LION-v3-Instruct \cite{sea_lion_2024}  & 0.847 & 0.787 & 0.793 & 0.733 & 0.790 \\ \cline{2-7}
& Qwen2.5-7B-Instruct \cite{yang2024qwen2technicalreport} & 0.847 & 0.84 & 0.753 & 0.713 & 0.788 \\ \cline{2-7}
& SeaLLMs-v3-7B-Chat \cite{zhang2024seallms3openfoundation} & 0.833 & 0.727 & 0.74 & 0.687 & 0.747 \\
\hline
\end{tabular}
\end{adjustbox}
\end{table}

\begin{table}[h]
\centering
\caption{Cross-LogiQA}
\label{tab:cross-logiqa}
\begin{adjustbox}{width=1.00\textwidth,center}
\begin{tabular}{|l|l|cccc|c|}
\hline
\textbf{Model Series} & \textbf{Model} & \textbf{English} & \textbf{Chinese} & \textbf{Indonesian} & \textbf{Malay} & \textbf{Avg } \\
\hline
\multirow{4}{*}{LLaMA3} & \textbf{Meta-Llama-3.1-8B-Instruct} & 0.585 & \textbf{0.585} & 0.455 & \textbf{0.523} & \textbf{0.537} \\ \cline{2-7}
& LLaMA-3-MERaLiON-8B-Instruct & 0.591 & 0.528 & \textbf{0.494} & 0.489 & 0.526 \\  \cline{2-7}
& Meta-Llama-3-8B-Instruct & \textbf{0.602} & 0.523 & 0.438 & 0.483 & 0.512 \\ \cline{2-7}
& Llama3-8B-SEA-LION-v2.1-Instruct & 0.528 & 0.517 & 0.403 & 0.443 & 0.473 \\
\hline
\multirow{4}{*}{Others} & \textbf{Qwen2.5-7B-Instruct} & \textbf{0.693} & \textbf{0.71} & \textbf{0.631} & 0.534 & \textbf{0.642} \\ \cline{2-7}
& Gemma-2-9B-IT & 0.659 & 0.636 & 0.585 & \textbf{0.602} & 0.621 \\ \cline{2-7}
& Gemma2-9B-SEA-LION-v3-Instruct & 0.636 & 0.642 & 0.557 & 0.551 & 0.597 \\ \cline{2-7}
& SeaLLMs-v3-7B-Chat & 0.568 & 0.585 & 0.494 & 0.517 & 0.541 \\
\hline
\end{tabular}
\end{adjustbox}
\end{table}

We conducted comprehensive evaluations using several benchmarks to assess the multilingual and instruction-following performance of MERaLiON. Key benchmarks include:

\begin{itemize}
    \item \textbf{Cross-MMLU} \cite{wang-etal-2024-seaeval} : A subset of the MMLU dataset translated into multiple languages, including English, Chinese, Indonesian, and Malay. It aims to measure the model’s ability to handle general knowledge queries across these diverse linguistic contexts.
    \item \textbf{Cross-LogiQA} \cite{wang-etal-2024-seaeval} : Building on the original LogiQA dataset, Cross-LogiQA introduces multilingual versions of logical reasoning tasks. It provides parallel question sets in English, Chinese, and Indonesian that are designed to maintain logical equivalence across these languages. 
    \item \textbf{IndoMMLU} \cite{koto2023largelanguagemodelspass} : A benchmark designed to assess general knowledge and language understanding of large language models for Indonesian, particularly for domains such as law, medicine, and social science \cite{koto2023largelanguagemodelspass}.
    \item \textbf{CN-Eval:} A selected subset of C-Eval \cite{huang2023cevalmultilevelmultidisciplinechinese} and CMMLU \cite{li2024cmmlumeasuringmassivemultitask} curated to specifically assess a model's knowledge about the Chinese language, culture and socio-political context. This subset provides a focused metric for assessing LLM knowledge of China.
\end{itemize}


\subsection{Results}

The results of our evaluation highlight several strengths of MERaLiON:

\begin{table}[h]
\centering
\caption{IndoMMLU}
\label{tab:model_comparison}
\begin{adjustbox}{width=0.75\textwidth,center}
\begin{tabular}{|l|l|c|}
\hline
\textbf{Model Series} & \textbf{Model} & \textbf{Accuracy} \\
\hline
\multirow{4}{*}{LLaMA Series} & \textbf{LLaMA-3-MERaLiON-8B-Instruct} & \textbf{0.576} \\
\cline{2-3}
 & Llama3-8B-SEA-LION-v2.1-Instruct & 0.560 \\
\cline{2-3}
 & Meta-Llama-3.1-8B-Instruct & 0.548 \\
\cline{2-3}
 & Meta-Llama-3-8B-Instruct & 0.521 \\
\hline
\multirow{5}{*}{Others} & \textbf{GPT4o-0513} & \textbf{0.760} \\
\cline{2-3}
 & Gemma2-9B-SEA-LION-v3-Instruct & 0.626 \\
\cline{2-3}
 & Gemma-2-9B-IT & 0.621 \\
\cline{2-3}
 & Qwen2.5-7B-Instruct & 0.582 \\
\cline{2-3}
 & SeaLLMs-v3-7B-Chat & 0.541 \\
\hline
\end{tabular}
\end{adjustbox}
\end{table}

\begin{table}[h]
\centering
\caption{CN-Eval}
\label{tab:cn_eval}
\begin{adjustbox}{width=0.75\textwidth,center}
\begin{tabular}{|l|l|l|}
\hline
\textbf{Model Series} & \textbf{Model} & \textbf{Accuracy} \\ \hline
\multirow{5}{*}{LLaMA Series} & \textbf{LLaMA-3-MERaLiON-8B-Instruct} & \textbf{0.514} \\ \cline{2-3}
 & Llama3-8B-SEA-LION-v2.1-Instruct & 0.505 \\ \cline{2-3}
 & Llama3-8B-SEA-LION-v2-Instruct & 0.495 \\ \cline{2-3}
 & Meta-Llama-3-8B-Instruct & 0.467 \\ \cline{2-3}
 & Meta-Llama-3.1-8B-Instruct & 0.457 \\ \hline
\multirow{5}{*}{Others} & \textbf{Qwen2-7B-Instruct} & \textbf{0.829} \\ \cline{2-3}
 & GPT4o-0513 & 0.810 \\ \cline{2-3}
 & Qwen2.5-7B-Instruct & 0.800 \\ \cline{2-3}
 & Gemma2-9B-SEA-LION-v3-Instruct & 0.590 \\ \cline{2-3}
 & Gemma-2-9B-IT & 0.581 \\ \hline
\end{tabular}
\end{adjustbox}
\end{table}

\begin{itemize}


    

    \item On Cross-MMLU and Cross-LogiQA, as shown in Table \ref{tab:cross-mmlu} and Table \ref{tab:cross-logiqa}, MERaLiON outperforms the baseline Meta-Llama-3.1-8B-Instruct model on Cross-MMLU, demonstrating superior general knowledge coverage in Chinese and Indonesian, while maintaining a strong, balanced performance in English. Similarly, MERaLiON improves logical reasoning in Indonesian, indicating that its continued pre-training and model merging strategies effectively enhance both factual knowledge and reasoning capabilities.

    \item On IndoMMLU, as reported in Table \ref{tab:model_comparison}, MERaLiON significantly outperforms the baseline Llama-3.1-8B-Instruct model, achieving 0.576 accuracy versus 0.548, highlighting its improved understanding of Indonesian language and domain-specific nuances.

    \item For CN-Eval, MERaLiON achieves 0.514 accuracy compared to 0.457 for Llama-3.1-8B-Instruct, as shown in Table \ref{tab:cn_eval}. This result demonstrates MERaLiON’s efficacy in retaining and enhancing knowledge related to China.
\end{itemize}

These benchmarks demonstrate the superior capability of MERaLiON-TextLLM to handle multilingual tasks and deliver improved cross-lingual understanding. Our training approach ensures the model is well-suited for diverse language tasks.

\section{Conclusion and Future Work}

The LLaMA-3-MERaLiON-8B-Instruct model improves multilingual NLP for Indonesian and Chinese, by addressing cross-lingual challenges with effective pretraining and model merging.

Evaluation results demonstrate MERaLiON-TextLLM's strengths, such as better reasoning and question-answering on Cross-MMLU and Cross-LogiQA through balanced multilingual training, strong accuracy on IndoMMLU and CN-Eval from effective dataset preparation, enhanced instruction-following via weight merging with minimal tuning, and resource efficiency through optimized TPU/GPU strategies for scalable training.

Future directions include:
\begin{itemize}
    \item Expanding to more underrepresented languages like Tagalog, Thai, and Vietnamese.
    \item Developing enhanced evaluation frameworks with human judgment and multimodal input.
    \item Applying the model to tasks such as translation, summarization, and content analytics.
\end{itemize}

MERaLiON demonstrates the potential for resource-efficient multilingual NLP, setting the foundation for broader linguistic coverage and consistent multilingual reasoning.

\section*{Acknowledgement}

We extend our sincere gratitude to Nattadaporn Lertcheva, Xi Wang, Kui Wu, Yang Ding, Nabilah Binte Md Johan for their invaluable contributions to data, insightful discussions, and future work explorations.

This research is supported by the National Research Foundation, Singapore and Infocomm Media Development Authority, Singapore under its National Large Language Models
Funding Initiative. Any opinions, findings and conclusions or
recommendations expressed in this material are those of the
author(s) and do not reflect the views of National Research
Foundation, Singapore and Infocomm Media Development
Authority, Singapore.

The resources and platforms provided by Singapore NSCC Aspire2A+ and The TPU Research Cloud. We thank all contributors and collaborators who have made this effort possible.


\bibliographystyle{abbrvnat}
\bibliography{template}

\end{document}